\documentclass[pdflatex]{nolta}
\usepackage{mathtools}
\usepackage{bm}

\Vol{17}%
\No{1}%
\Year{2026}%
\Month{1}%
\PaperID{2026ENP0886}

\title{Geometric and dynamical analysis of attractor boundaries and storage limits in kernel Hopfield networks}

\AUTHOR{
\author{Akira Tamamori}{1}\orcid{0009-0000-8893-0058}
}

\AFFILIATE{ \affiliate{Faculty of Information Science, Aichi
    Institute of Technology\\1247 Yachigusa, Yakusa-cho, Toyota-shi, Aichi 470-0392, Japan}{1} }

\received{10}{27}{20XX}
\revised{12}{29}{20XX}
\published{7}{1}{20XX}

\begin{document}

\begin{abstract}
  High-capacity associative memories based on Kernel Logistic
  Regression (KLR) exhibit strong storage capabilities, but the
  dynamical and geometric mechanisms underlying their stability remain
  poorly understood.  This paper investigates the global geometry of
  attractor basins and the mechanisms governing the storage limit in
  KLR-trained Hopfield networks. We combine empirical evaluations
  using random sequences and real-world image embeddings (CIFAR-10)
  with morphing experiments and statistical Signal-to-Noise Ratio
  (SNR) analysis.  Our experiments show that the network achieves a
  storage capacity for random sequences up to $P/N \approx 16$, while
  maintaining stable retrieval for structured data at effective loads
  near $P/N \approx 20$. Morphing analysis indicates that attractors
  on the ``Ridge of Optimization'' are separated by sharp,
  phase-transition-like boundaries, characterized by steep effective
  potential barriers and critical slowing down.  Furthermore, by
  comparing an SNR analysis with a geometric reference point inspired
  by Cover's theorem, we show that the practical storage limit is
  governed primarily not by a lack of geometric separability in the
  feature space, but by the loss of dynamical stability against
  crosstalk noise. These findings suggest that KLR networks function
  as highly localized exemplar-based memories that operate near the
  onset of dynamical collapse, providing a useful perspective on the
  design of robust, large-scale retrieval systems.
\end{abstract}
\begin{keywords}
  kernel Hopfield network, associative memory, storage capacity,
  attractor geometry, signal-to-noise ratio, exemplar-based memory
\end{keywords}

\maketitle

\section{Introduction}
\label{sec:intro}
Associative memory, the ability to retrieve complete data patterns
from partial or noisy cues, is a fundamental mechanism in both
biological and artificial neural systems. The Hopfield
network~\cite{Hopfield1982} provides a canonical model for this
process, where memories are stored as stable fixed points (attractors)
of an energy landscape. While the classical model is theoretically
elegant, it suffers from a severe storage capacity limit for random
patterns ($P\approx 0.14N$)~\cite{Amit1985}, thereby limiting its
practical applicability.

Recent research has explored two primary avenues to overcome this
limitation. The first, exemplified by Modern Hopfield
Networks~\cite{Krotov2016, Ramsauer2021}, redesigns the energy
function to achieve higher capacity. The second approach, which we
pursue here, retains the standard quadratic energy structure (in a
feature space) but employs advanced learning algorithms. Our prior
work showed that Kernel Logistic Regression (KLR) learning falls into
this category, achieving substantially higher storage capacity for
random patterns ($P/N > 4.0$)~\cite{tamamori_letter,tamamori_nolta_a,
  tamamori_nolta_b}. These studies also identified a specific
hyperparameter regime, termed the ``Ridge of Optimization'', where
attractor stability is maximized, especially under high-load
conditions.

Despite these advances, the underlying dynamical mechanisms remain
insufficiently understood. The storage capacity for temporal sequences
and the performance on correlated, real-world data have not been
systematically investigated. Furthermore, the geometric principles
that govern the global structure of the attractor landscape, such as
the nature of the boundaries between basins and the physical origin of
the ultimate capacity limit, require deeper analysis.

This paper addresses these gaps through a comprehensive investigation
into the geometric and dynamical properties of KLR Hopfield networks
on the Ridge. We extend the model's application to sequence learning
and real-world image storage, and by combining empirical observations
with theoretical analyses, we elucidate the principles governing its
high performance. Our main contributions are:

\begin{enumerate}
\item We demonstrate that the KLR network achieves an high storage
  capacity for random sequences ($P/N\approx 16$) and exhibits strong
  performance on structured data (CIFAR-10 embeddings) at even higher
  loads ($P/N\approx 20$), reflecting the low-dimensional structure of
  the data.
  
\item We characterize the global geometry of the attractor
  basins. Through morphing experiments, we show that attractors are
  separated by sharp boundaries reminiscent of phase transitions. We
  visualize the effective potential landscape and identify a critical
  slowing down phenomenon, providing insight into the geometry of the
  separatrix.

\item We provide a theoretical explanation for the memory collapse
  mechanism. By contrasting a geometric separability analysis (related
  to Cover's theorem~\cite{Cover1965}) with a statistical
  Signal-to-Noise Ratio (SNR) analysis, we suggest that the practical
  capacity is limited not by geometric constraints, but by the
  dynamical stability of the signal against crosstalk noise.
\end{enumerate}

The remainder of this paper is organized as follows.
Section~\ref{sec:methods} outlines the network model and experimental
datasets.  Section~\ref{sec:phenomena} presents empirical results on
sequence and real-world data storage.  Section~\ref{sec:geometry}
analyzes the geometry of attractor boundaries.
Section~\ref{sec:mechanism} provides a theoretical analysis of the
capacity limit.  Section~\ref{sec:discussion} discusses the broader
implications of the findings, and Section~\ref{sec:conclusion}
concludes the paper.

\section{Methods}
\label{sec:methods}
In this section, we define the kernel Hopfield network
model~\cite{tamamori_letter, tamamori_nolta_a} and its extension to
sequence memory. We also describe the synthetic and real-world
datasets used in our experiments.

\subsection{Kernel Hopfield Network}
We consider a network of $N$ neurons with state
$\bm{s} \in \{-1, 1\}^{N}$.  The dynamics are defined through a
kernel-induced energy landscape, where the local field $h_{i}(\bm{s})$
for neuron $i$ is given by:
\begin{equation}
h_i(\bm{s}) = \sum_{\mu=1}^P \alpha_{\mu i} K(\bm{s}, \boldsymbol{\xi}^\mu),
\end{equation}
where $\{\boldsymbol{\xi}^\mu\}_{\mu=1}^{P}$ are the stored patterns,
$\alpha_{\mu i}$ are the dual variables learned via Kernel Logistic
Regression (KLR)~\cite{Schölkopf2001}, and $K(\cdot, \cdot)$ is the
kernel function. In this study, we use the RBF kernel,
$K(\bm{x}, \bm{y}) = \exp(-\gamma\|\bm{x} - \bm{y}\|^{2} )$, with the
locality parameter $\gamma$.

Unless otherwise specified, we set the kernel locality parameter to
$\gamma=0.02$, a value corresponding to the ``Ridge'' regime
identified in our previous
analysis~\cite{tamamori_nolta_b}. Furthermore, when analyzing behavior
near the storage limit of static memory, we typically use a high
storage load of $P/N = 10.0$ (i.e., $P=1000, N=100$). Exceptions to
these settings, such as experiments using real-world datasets, are
noted where appropriate.

To characterize the stability landscape, we define a pseudo-energy
function, $V(\bm{s})$, which measures the alignment between the
current state $\bm{s}$ and its local field $\bm{h}(\bm{s})$:
\begin{equation}
  V(\bm{s}) \coloneqq -\sum_{i=1}^N s_i h_i(\bm{s}) = -\sum_{i=1}^N s_i \sum_{\mu=1}^P \alpha_{\mu i} K(\bm{s}, \boldsymbol{\xi}^\mu).
  \label{eq:lyapunov}
\end{equation}
A lower value of $V(\bm{s})$ indicates a stronger alignment and thus a
locally more stable configuration. Since the synchronous update rule
does not guarantee monotonic energy decrease, $V(\bm{s})$ is used here
as a heuristic measure of stability, rather than a strict Lyapunov
function. We use this measure to examine the effective potential
landscape in Section~\ref{sec:geometry}.

\subsection{Extension to Sequence Memory}
To store a sequence of patterns
$\boldsymbol{\xi}^1 \rightarrow \boldsymbol{\xi}^2 \rightarrow \cdots
\rightarrow \boldsymbol{\xi}^P \rightarrow \boldsymbol{\xi}^1$, we
extend the KLR framework to hetero-associative memory. Unlike the
conventional auto-associative setting where the desired output
corresponding to input $\boldsymbol{\xi}^\mu$ is
$\boldsymbol{\xi}^\mu$ itself, we train the network to map the current
pattern $\boldsymbol{\xi}^\mu$ to the next pattern in the sequence,
$\boldsymbol{\xi}^{\mu + 1}$ (with
$\boldsymbol{\xi}^{P+1} = \boldsymbol{\xi}^{1}$). This introduces
asymmetry into the implicit weight matrix, enabling the network to
generate a limit-cycle trajectory that successively traverses the
stored patterns in the correct order. The update dynamics remain
unchanged: $s_{i}(t+1) = \text{sign}(h_{i}(\bm{s}(t)))$.

\subsection{Datasets}

We evaluate the model using two types of datasets to assess
its storage capacity and robustness under different conditions.

\paragraph{Random Patterns}
For theoretical analysis and evaluation near the storage limit,
we use random binary patterns in which each element
$\xi_i^\mu$ is independently drawn from $\{-1,1\}$ with equal
probability.
This standard setting allows direct comparison with classical
capacity theories.

\paragraph{Image Embeddings (CIFAR-10)}
To evaluate performance on structured data, we use the CIFAR-10
dataset~\cite{Krizhevsky2009}.  Rather than using raw pixel values, we
extract feature embeddings from the 10,000 images in the standard test
set using a pre-trained ResNet-18 model~\cite{He2016}.  The
512-dimensional feature vectors are centered and binarized using the
sign function to obtain patterns
$\boldsymbol{\xi}^\mu \in \{-1,1\}^{512}$.  This approach enables the
network to exploit the semantic structure of natural images while
maintaining binary state representations.  For experiments with fewer
than 10,000 patterns, we use randomly selected subsets of the test
set.

\subsection{Experimental Protocol for Boundary Analysis}
To investigate the geometry of attractor basin boundaries, we employ a
morphing procedure based on linear interpolation.  Let
$\boldsymbol{\xi}^{A}$ and $\boldsymbol{\xi}^{B}$ be two distinct
stored patterns.  We construct an intermediate state $\bm{s}(r)$
parameterized by the interpolation ratio $r \in [0,1]$ as follows:
\begin{equation}
  \bm{s}(r) = \mathrm{sign}\left( (1-r) \boldsymbol{\xi}^A + r \boldsymbol{\xi}^B + \boldsymbol{\epsilon} \right),
\label{eq:morphing}
\end{equation}
where $\boldsymbol{\epsilon} \sim \mathcal{N}(\bm{0}, \nu^{2}\bm{I})$
is a small noise term added to break symmetry at the midpoint
$(r=0.5)$. In our experiments, we set the noise standard deviation to
$\nu = 0.01$. Using $\bm{s}(r)$ as the initial state, we iterate the
retrieval dynamics until convergence and measure the overlap with
$\boldsymbol{\xi}^{A}$ and $\boldsymbol{\xi}^{B}$.

Furthermore, to visualize the energy landscape along this path, we
define the Effective Potential $U(r)$ as the value of the
pseudo-energy function $V(\bm{s})$ (defined in
Eq.~(\ref{eq:lyapunov})) evaluated at the normalized continuous state:
\begin{equation}
  U(r) \coloneqq V(\tilde{\bm{s}}(r)) = -\sum_{i=1}^N \tilde{s}_i(r) h_i(\tilde{\bm{s}}(r)),
\label{eq:effective_potential}
\end{equation} 
where
$\tilde{\bm{s}}(r) = \sqrt{N} \frac{(1-r)\boldsymbol{\xi}^{A} + r
  \boldsymbol{\xi}^{B}}{\|(1-r)\boldsymbol{\xi}^{A} + r
  \boldsymbol{\xi}^{B}\|}$ is the interpolation vector projected onto
the hypersphere of radius $\sqrt{N}$. While the actual network
dynamics operate in a discrete state space, this continuous relaxation
serves as a heuristic approximation of the effective transition
landscape for state evolution.  It allows us to probe the potential
barrier separating the two attractors and provides a qualitative view
of the mechanisms governing the restoration dynamics.

\subsection{Experimental Protocol for Sequence Memory Analysis}
\label{sec:sequence_protocol}
To evaluate the storage capacity for temporal sequences, we conducted
numerical simulations using the following protocol.  For each sequence
length $P$, we performed 10 trials with independently generated random
patterns.  In each trial, we trained the network and then iterated the
network dynamics starting from the first pattern
($\bm{s}(0)=\boldsymbol{\xi}^{1}$) for a total of $6P$ update steps.

Recall is considered successful if the network correctly reproduces
the entire sequence for at least one full cycle ($P$ steps) without
any transition error.  An error is defined as the state vector
$\bm{s}(t)$ differing from the target pattern
$\boldsymbol{\xi}^{t+1}$.  Memory collapse is identified as the point
at which the network fails to satisfy this criterion, either by
converging to a fixed point or by entering a trajectory deviating from
the target sequence.  The capacity limit $P_c$ is defined as the
maximum sequence length for which recall is successful across all
trials.

\section{Empirical Phenomena Beyond the Classical Capacity Limit}
\label{sec:phenomena}

In this section, we present empirical results on the storage capacity
of KLR-trained Hopfield networks, demonstrating storage capacities
that substantially exceed classical theoretical bounds.

\subsection{Sequence Memory Capacity}

\begin{figure*}[t]
  \centering
  \begin{minipage}{0.90\hsize}
    \centering
    \includegraphics[width=\linewidth]{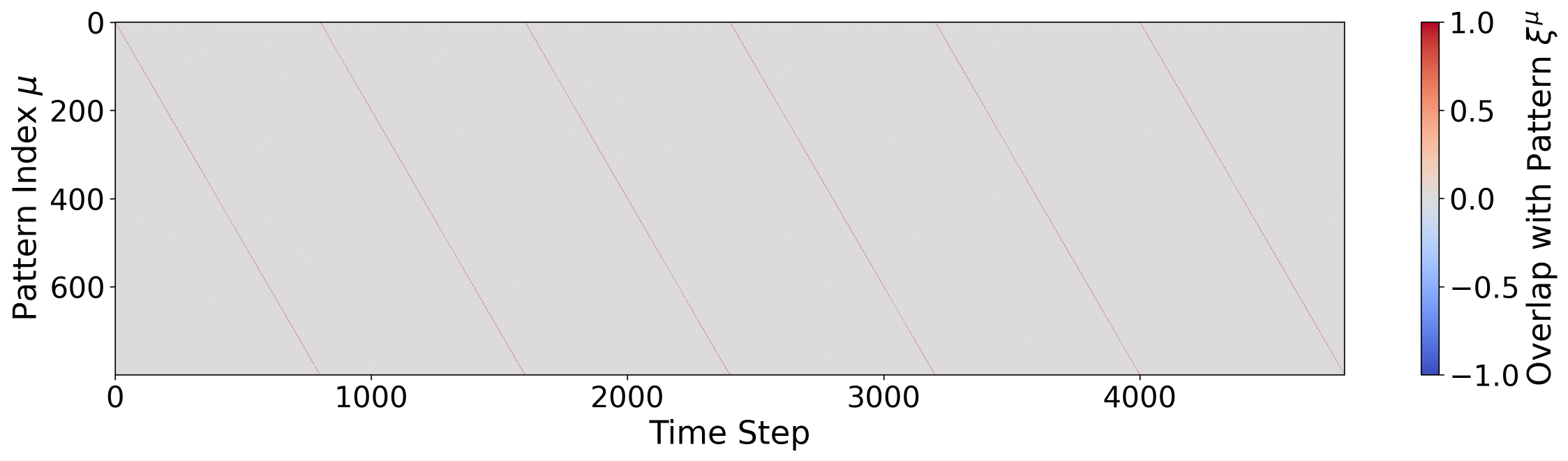}
    \centerline{(a) Success case ($P=800$)}
  \end{minipage}
  \begin{minipage}{0.90\hsize}
    \centering
    \includegraphics[width=\linewidth]{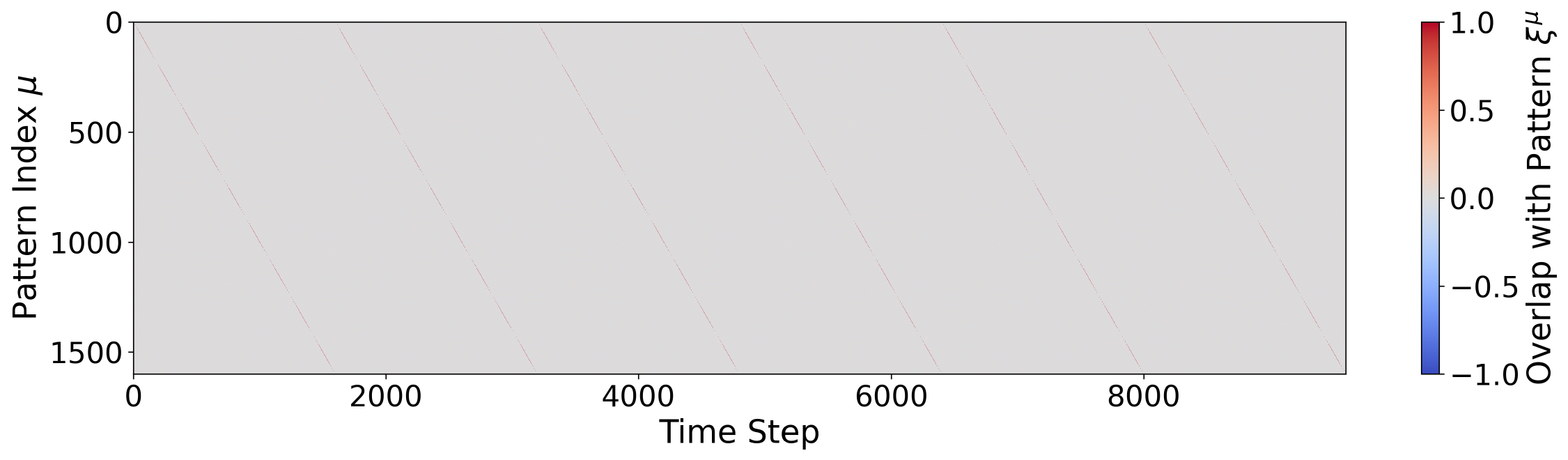}
    \centerline{(b) Success case ($P=1600$)}
  \end{minipage}
  \begin{minipage}{0.90\hsize}
    \centering
    \includegraphics[width=\linewidth]{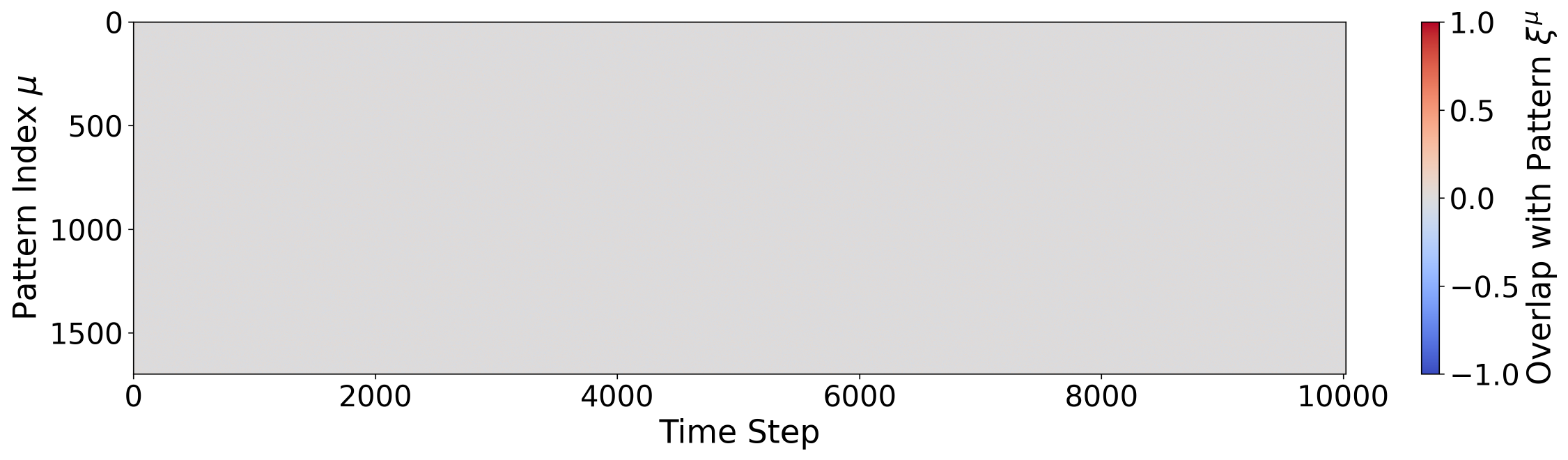}
    \centerline{(c) Failure case ($P=1700$)}
  \end{minipage}
  \caption{Sequence recall dynamics at different loads.  Time evolution
    of the average overlap between the network state and stored patterns
    across 10 independent trials ($N=100$).  The heatmap shows the
    average overlap value (red: $+1$, blue: $-1$).  (a) Success case
    ($P=800$): The network robustly recalls the sequence.  (b)
    Near-capacity case ($P=1600$): Recall remains successful, although
    the overlap signal becomes weaker.  (c) Failure case ($P=1700$): The
    retrieval dynamics collapse and fail to reproduce the sequence.}
  \label{fig:sequence_dynamics}
\end{figure*}

We first investigated the storage capacity for temporal sequences,
following the protocol detailed in
Section~\ref{sec:sequence_protocol}. We trained the network to recall
a cyclic sequence of random patterns with increasing length~$P$,
averaging the results over 10 independent trials.

Figure~\ref{fig:sequence_dynamics} shows representative recall
dynamics for different storage loads.  At a moderate load of $P=800$
($P/N=8.0$), the network successfully traverses the sequence with high
overlap (Fig.~\ref{fig:sequence_dynamics}(a)).  Even at $P=1600$
($P/N=16.0$), the network maintains a stable, albeit weaker,
limit-cycle (Fig.~\ref{fig:sequence_dynamics}(b)).  However,
increasing the load slightly further to $P=1700$ causes the dynamics
to collapse, preventing successful sequence retrieval
(Fig.~\ref{fig:sequence_dynamics}(c)).

Two observations emerge from these experiments.  First, the storage
capacity for random sequences in KLR networks reaches
$P/N \approx 16$, which is more than two orders of magnitude higher
than the classical Hopfield limit of $P/N \approx 0.14$.  Second, the
transition from successful recall to failure is sharp, a behavior
reminiscent of a phase transition.

\subsection{Storage of Real-World Data}

Next, we evaluated the storage performance of the network using
structured patterns derived from CIFAR-10 embeddings ($N=512$).  As
detailed in Section~\ref{sec:methods}, we trained the network in an
auto-associative configuration (static memory) using KLR to store
progressively larger sets of image patterns.  The training was
performed for 500 iterations using a learning rate of $0.1$ and a
weight decay of $0.01$.

To evaluate robustness against input noise, all stored patterns were
corrupted with random bit flips and then used as initial states to
test whether the network could retrieve the original clean patterns.
To ensure statistical reliability, all reported accuracies were
averaged over 10 independent trials.  For $P < 10{,}000$, a new subset
of images was randomly sampled for each trial; for $P=10{,}000$, the
noise realizations were randomized across trials.

\begin{table}[t]
\centering
\caption{Storage capacity and noise robustness for CIFAR-10 embeddings
  ($N=512$).  The network maintains perfect recall accuracy even at a
  load of $P/N \approx 19.5$.  Accuracies are reported for both
  noiseless inputs and inputs corrupted with 10\% random bit flips,
  averaged over 10 independent trials.}
\label{tab:cifar_capacity}
\setlength{\tabcolsep}{0.5\tabcolsep}
\begin{tabular*}{\columnwidth}{@{\extracolsep{\fill}}cccc}
\hline
\textbf{Patterns} & \textbf{Load} & \textbf{Acc. (0\%)} & \textbf{Acc. (10\%)} \\
($P$) & ($P/N$) & (Noise) & (Noise) \\
\hline
1,000 & 1.95 & 1.0000 & 1.0000 \\
3,000 & 5.86 & 1.0000 & 1.0000 \\
5,000 & 9.77 & 1.0000 & 1.0000 \\
8,000 & 15.63 & 1.0000 & 1.0000 \\
\textbf{10,000} & \textbf{19.53} & \textbf{1.0000} & \textbf{1.0000} \\
\hline
\end{tabular*}
\end{table}

Table~\ref{tab:cifar_capacity} summarizes the results.  The network
successfully stored and retrieved all 10,000 patterns from the
CIFAR-10 test set with essentially perfect bit-wise accuracy (e.g.,
$1.0000 \pm 0.0000$), even in the presence of 10\% random input noise.
This corresponds to an effective storage load of $P/N \approx 19.5$.

These results suggest that the KLR framework can exploit the
structural correlations inherent in real-world data embeddings.
Unlike random patterns, which are approximately orthogonal and largely
unstructured, real-world data typically lies on a lower-dimensional
manifold.  The kernel mapping enables the network to separate these
correlated patterns with high precision, resulting in an effective
storage load substantially higher than that observed for random
patterns.

\section{The Geometry of Attractor Boundaries}
\label{sec:geometry}

\begin{figure*}[t]
  \centering
  \begin{minipage}{0.48\textwidth}
    \centering
    \includegraphics[width=\linewidth]{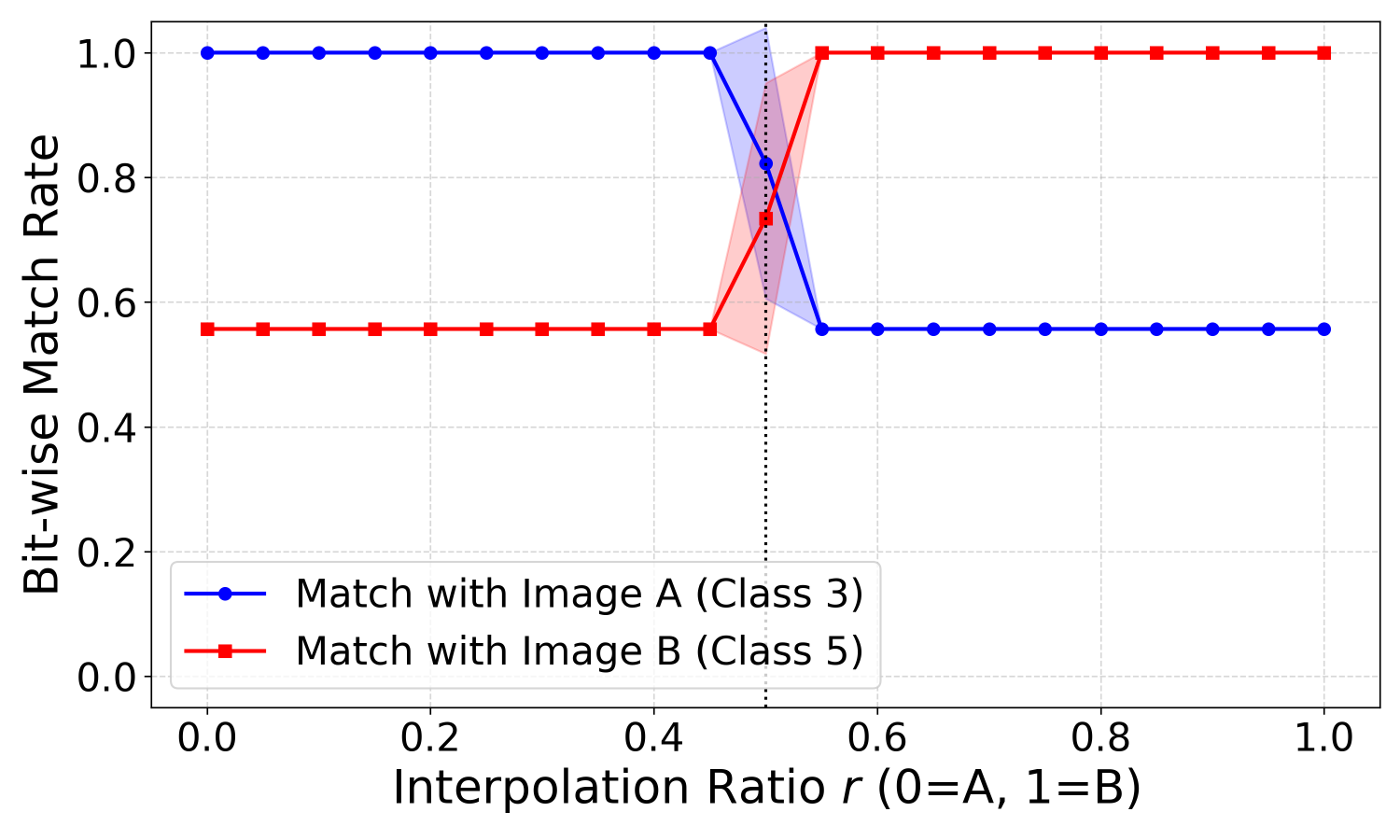}
    \includegraphics[width=\linewidth]{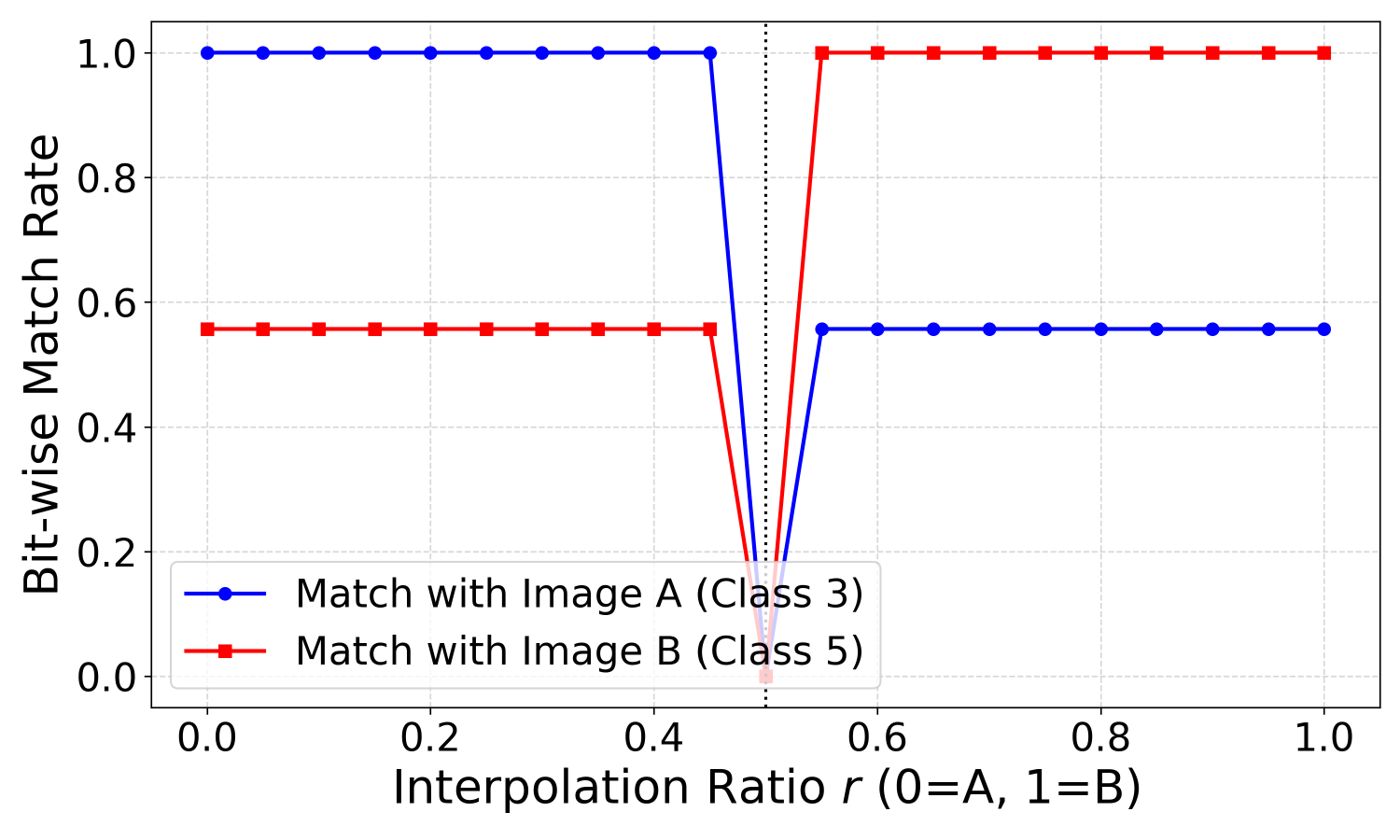}
    (a) Inter-class Morphing
  \end{minipage}
  \hfill
  \begin{minipage}{0.48\textwidth}
    \centering
    \includegraphics[width=\linewidth]{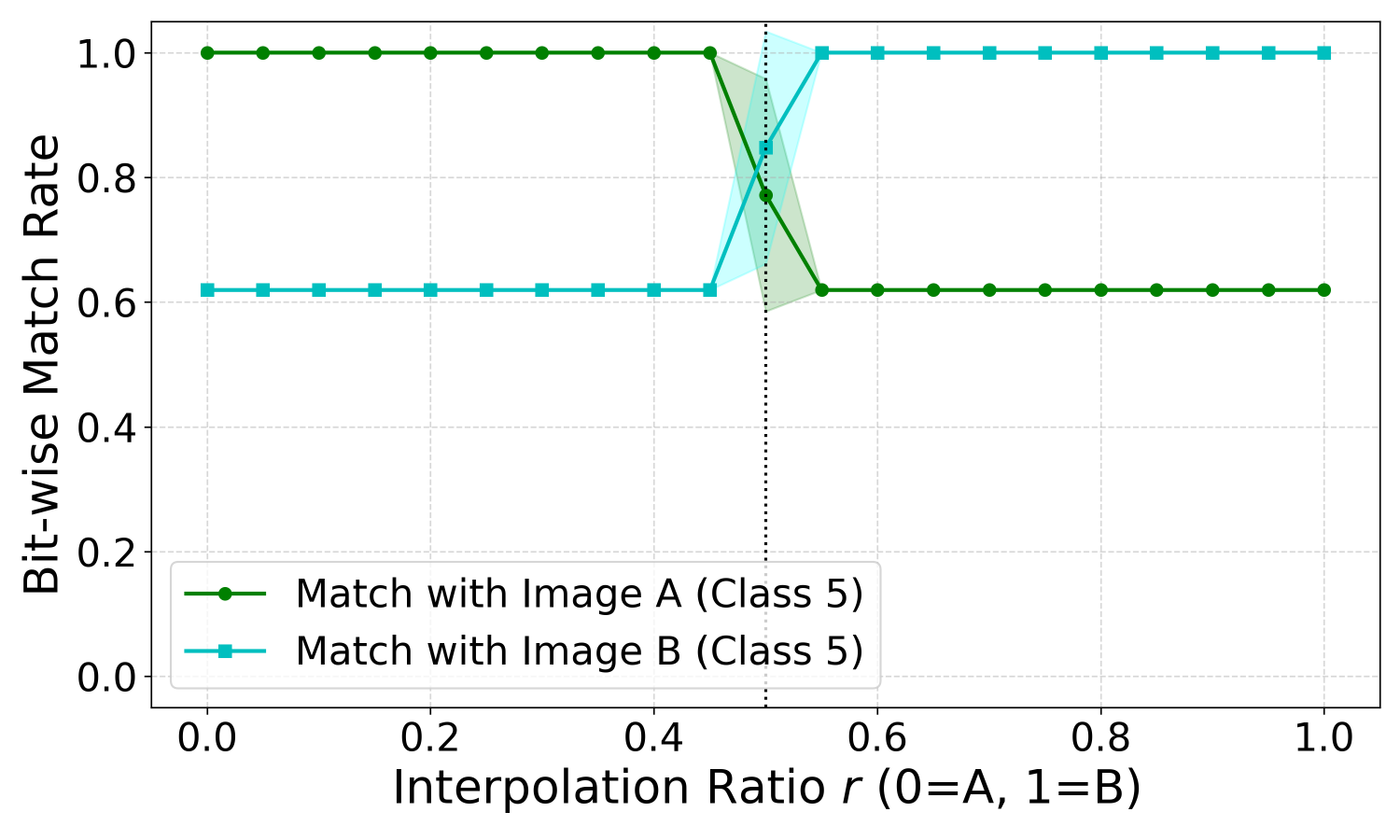}
    \includegraphics[width=\linewidth]{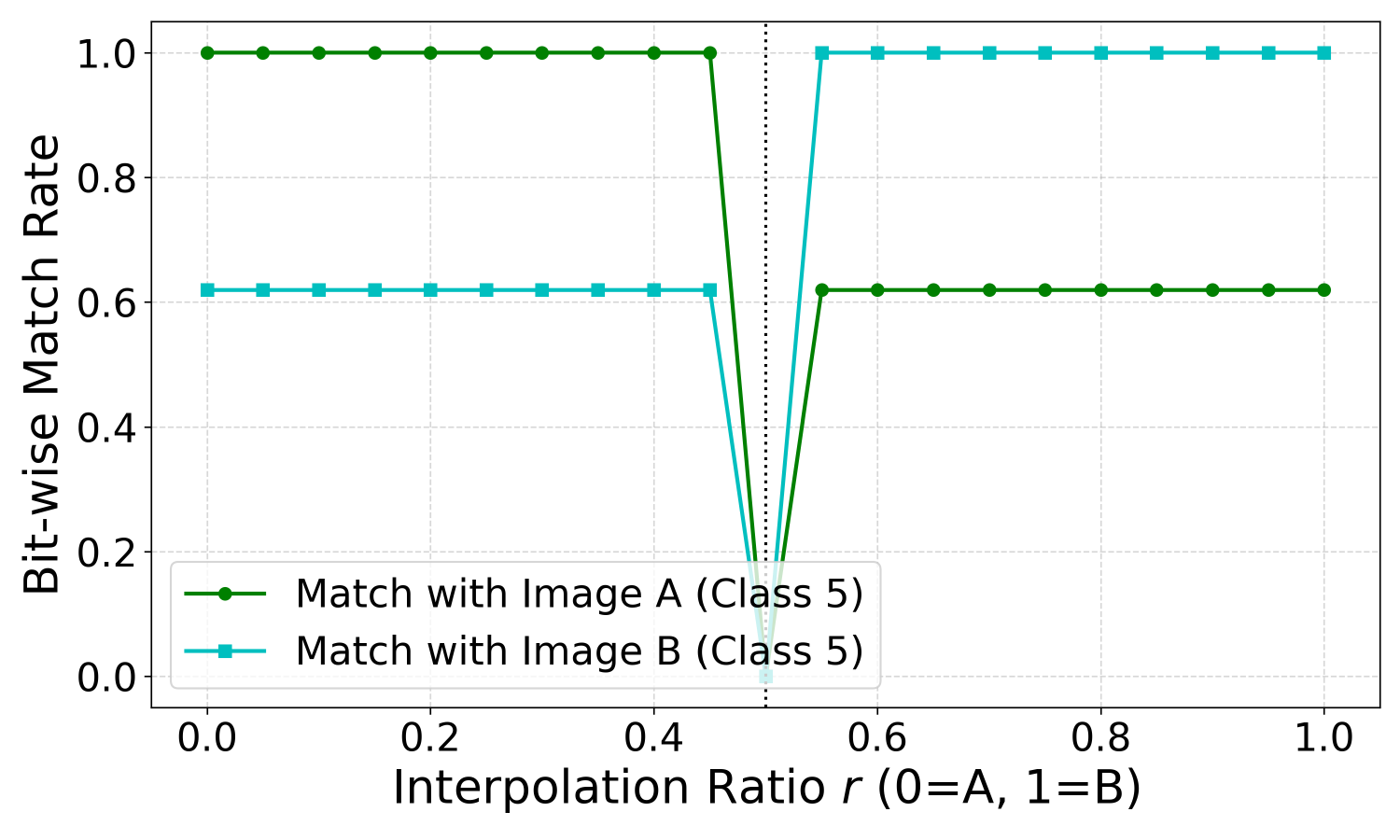}
    (b) Intra-class Morphing
  \end{minipage}
  \caption{Attractor transition analysis.  Comparison of recall
    dynamics in the Ridge regime ($\gamma=0.02$, top row) and the
    Local regime ($\gamma=5.0$, bottom row).  Shaded areas indicate
    standard deviations over 10 trials.  (a) Inter-class morphing: The
    Ridge regime exhibits a sharp transition, whereas the Local regime
    converges to spurious states near the boundary.  (b) Intra-class
    morphing: A similarly sharp transition is observed in the Ridge
    regime, demonstrating robust separation even for semantically
    similar patterns.}
  \label{fig:attractor_transition}
\end{figure*}

To investigate the mechanism underlying the observed capacity and
robustness, we investigate the geometric structure of the attractor
basins, with a focus on the boundaries between attractor basins.

\subsection{Sharp Transitions Across Attractor Boundaries}
We examined the recall dynamics at the boundary between two attractors
by morphing between two stored patterns, $\boldsymbol{\xi}^{A}$ and
$\boldsymbol{\xi}^{B}$, using an interpolation parameter
$r \in [0, 1]$. We considered two scenarios using CIFAR-10 embeddings:
\textit{Inter-class} morphing (between distinct object classes, e.g.,
cat vs. dog) and \textit{Intra-class} morphing (between different
instances of the same class).

Figure~\ref{fig:attractor_transition} summarizes the results, averaged
over 10 independent trials with small random noise added to the
interpolated states. On the Ridge ($\gamma=0.02$, top row), both
Inter-class (a) and Intra-class (b) transitions exhibit a sharp,
step-like behavior. The network converges reliably to either
$\boldsymbol{\xi}^{A}$ or $\boldsymbol{\xi}^{B}$ with no intermediate
states, even when the input is very close to the midpoint
($r \approx 0.5$). This indicates that the transition region is highly
localized and the attractor basins are closely separated without
intermediate unstable regions, regardless of the semantic similarity
of the patterns.

In contrast, in the Local regime ($\gamma=5.0$, bottom row), the sharp
transition disappears in both cases.  Near the midpoint, the network
fails to recall either pattern and instead converges to a spurious
state with low overlap.  This confirms that a large $\gamma$ creates
isolated basins separated by unstable or spurious regions, preventing
the system from reliably selecting either attractor near ambiguous
inputs.

\subsection{Effective Potential Along the Morphing Path}
\begin{figure}[t]
  \centering
  \includegraphics[width=\columnwidth]{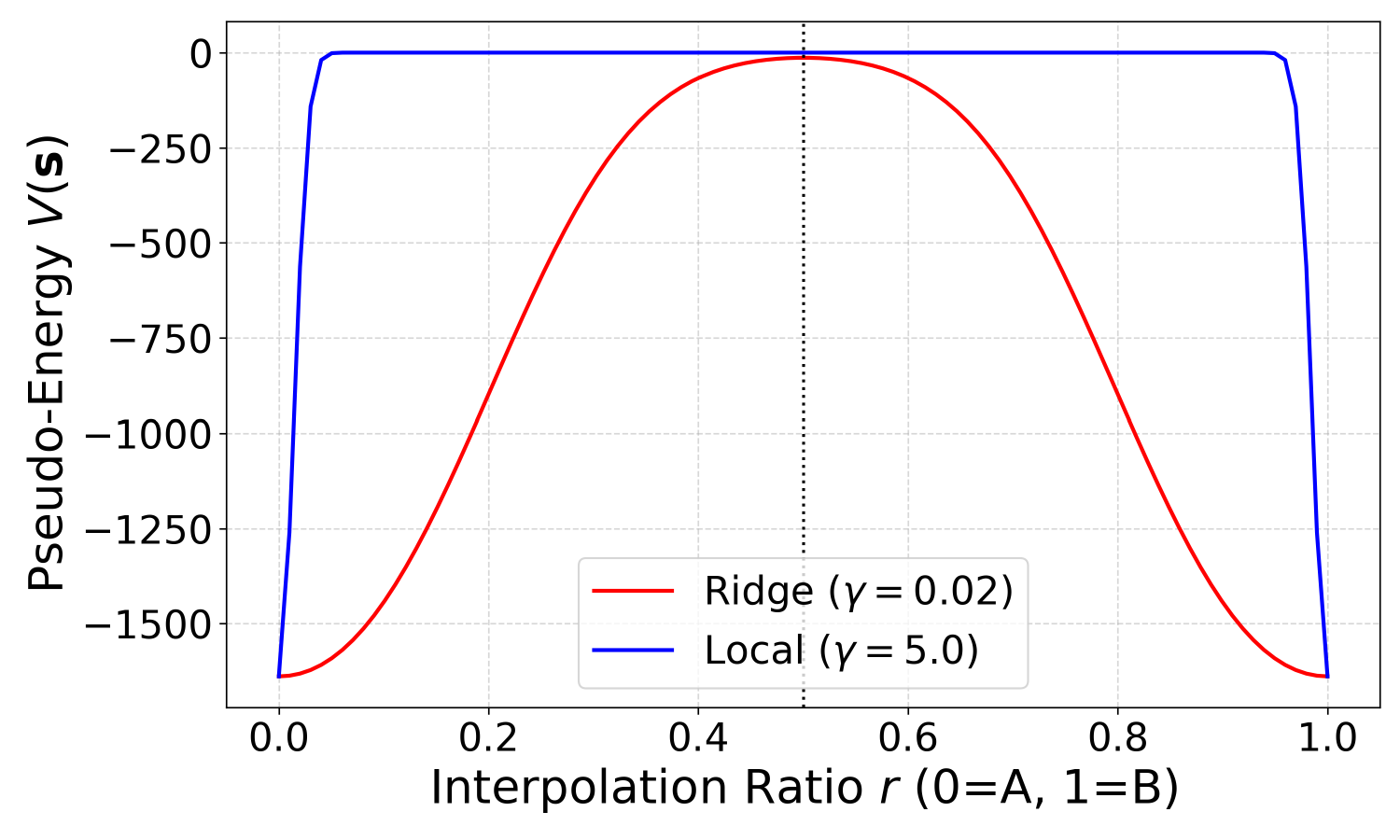}
  \caption{Effective potential along the morphing path.  The plot
    shows the heuristic pseudo-energy $V(\bm{s})$ evaluated along the
    normalized continuous interpolation path between two stored
    patterns.  The Ridge model (red) exhibits a steep potential
    barrier, whereas the Local model (blue) forms a relatively flat
    plateau.}
  \label{fig:effective_potential}
\end{figure}

The sharp transitions observed in the Ridge regime suggest a distinct
effective potential structure.  As defined in Eq.~(\ref{eq:lyapunov}),
we use the pseudo-energy $V(\bm{s})$ as a heuristic measure of
alignment, since synchronous updates do not guarantee monotonic energy
decrease.  We examined the effective potential $U(r)$ along the
morphing path (Fig.~\ref{fig:effective_potential}).

In the Ridge regime (red), the potential forms a pronounced
double-well-like structure separated by a steep barrier.  The
pseudo-energy increases smoothly but steeply from the attractors
toward the midpoint ($r=0.5$), creating a strong local gradient that
drives intermediate states toward the nearest basin.  In contrast, the
Local regime (blue) exhibits a wide, relatively flat plateau between
the attractors.  The lack of a significant potential gradient in this
region explains the emergence of spurious states and the failure to
reliably converge to either attractor near ambiguous inputs.

\subsection{Critical Slowing Down}
\begin{figure}[t]
  \centering
  \includegraphics[width=\columnwidth]{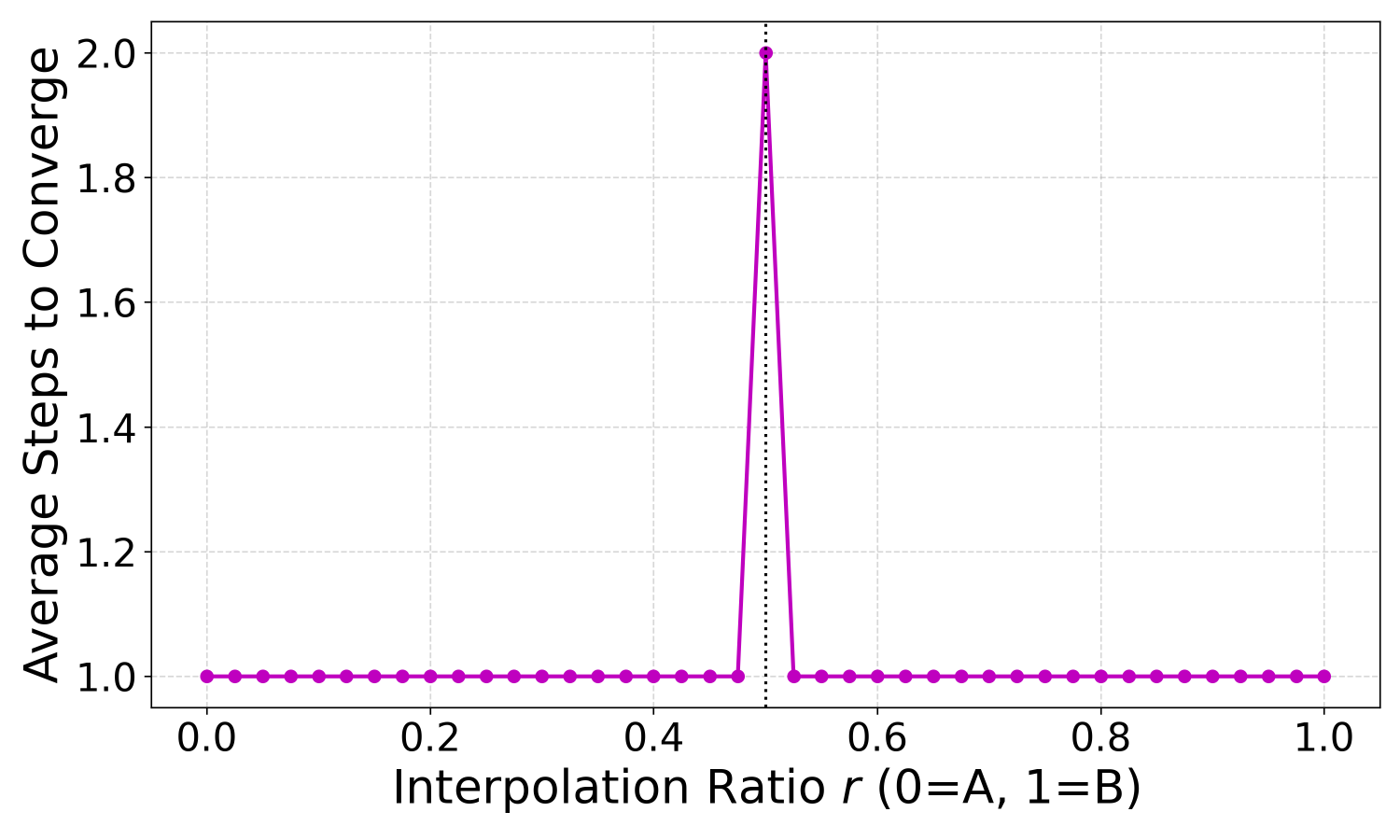}
  \caption{Critical slowing down near the attractor boundary.
    The plot shows the average number of update steps
    required for convergence (averaged over 10 trials)
    as a function of the interpolation ratio $r$
    between two patterns from the same class.}
  \label{fig:slowing_down}
\end{figure}

We further examined the dynamics near the boundary by measuring the
convergence time.  As shown in Fig.~\ref{fig:slowing_down}, the
convergence time in the Ridge regime remains minimal (1 step) over
most of the interpolation range, but exhibits a sharp peak exactly at
the boundary ($r=0.5$).

This phenomenon, reminiscent of \textit{Critical Slowing Down}
observed near phase transitions~\cite{Scheffer2009}, supports the
interpretation that the boundary acts as a dynamical separatrix.  The
narrow width of the peak further supports the steep structure of the
effective potential, where the unstable equilibrium is confined to a
narrow region near the boundary.

\section{Theoretical Mechanism of Memory Collapse}
\label{sec:mechanism}

We have established that KLR networks possess high storage capacity
for random sequences ($P/N \approx 16$) and strong performance on
structured data ($P/N \approx 20$), alongside sharp attractor
boundaries.  However, memory recall eventually collapses.  What
determines this capacity limit?  To investigate this question, we
contrast a statistical Signal-to-Noise Ratio (SNR) analysis with
geometric separability considerations.

\subsection{SNR Analysis of the Storage Limit}

We first analyze the stability of memory patterns using a statistical
mechanics approach.  The local field input
$h_{i}(\boldsymbol{\xi}^{\mu})$ to a neuron can be decomposed into a
signal term (from the target pattern $\boldsymbol{\xi}^{\mu}$) and a
crosstalk noise term (from all other patterns
$\boldsymbol{\xi}^{\nu\neq \mu}$).  We measured the mean signal
strength $S$ and the noise standard deviation $\sigma$ as functions of
the number of stored patterns~$P$.

\begin{figure}[t]
  \centering
  \includegraphics[width=\columnwidth]{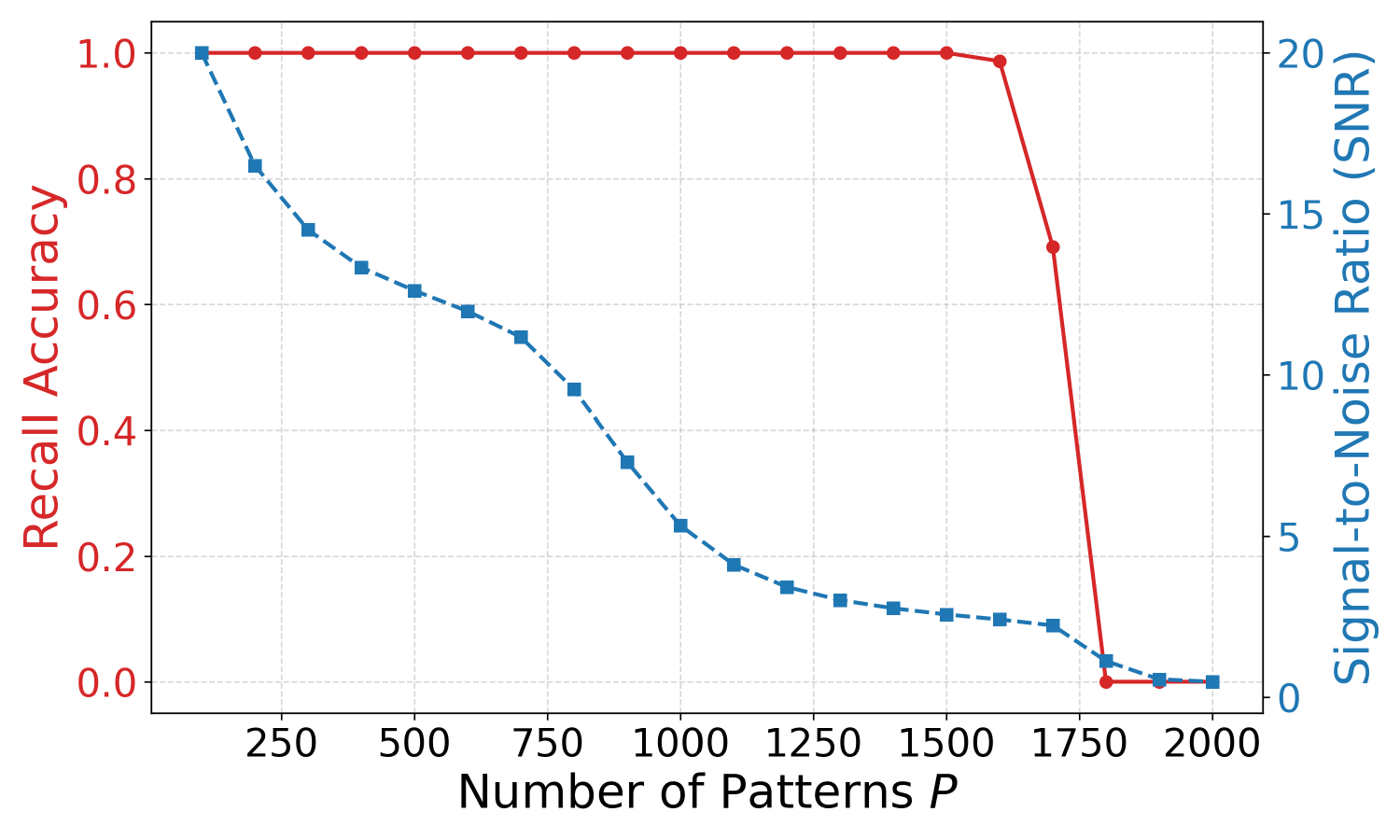}
  \caption{SNR analysis of the storage limit.  The recall accuracy
    (red) for random sequences collapses sharply when the
    signal-to-noise ratio (SNR, blue) drops below a threshold
    ($\approx 2.0$).}
  \label{fig:snr_analysis}
\end{figure}

Figure~\ref{fig:snr_analysis} plots the SNR ($S/\sigma$) alongside the
recall accuracy.  As $P$ increases, the SNR decreases monotonically
due to accumulating crosstalk noise.  The collapse of memory recall
($P \approx 1700$) closely coincides with the point at which the SNR
drops below a threshold (SNR $\approx 2.0$).  This threshold is
consistent with statistical intuition: assuming the crosstalk noise
follows a Gaussian distribution, an SNR of 2.0 implies that the signal
strength $S$ is approximately $2\sigma$.  Below this level, the
probability of bit-flip errors during a single update step becomes
sufficiently large for errors to accumulate iteratively, leading to a
global breakdown of stable recall.  This suggests that the storage
limit corresponds to a dynamical phase transition driven by the loss
of signal dominance over noise, consistent with classical Hopfield
theory but at a much higher scale.

\subsection{Geometric Separability vs. Dynamic Stability}
Is this limit determined primarily by the geometry of the feature
space?  Cover's theorem~\cite{Cover1965} states that $P$ random
patterns are linearly separable in $D$ dimensions with high
probability if $P < 2D$.  Although this is not a strict bound for
associative memory capacity, it provides a useful geometric reference
point for pattern separability.  In our kernel model, the feature
space is infinite-dimensional, so we use the effective dimension
$D_{\text{eff}}(\bm{K})$ of the kernel Gram matrix $\bm{K}$ as a proxy
for $D$.  We define this effective dimension using the participation
ratio~\cite{Litwin-Kumar2017, Susman2021} of the eigenvalues
$\{\lambda_k\}_{k=1}^P$ of $\bm{K}$:
\begin{equation}
  D_{\text{eff}}(\bm{K})
  =
  \frac{
    \left( \sum_{k=1}^P \lambda_k \right)^2
  }{
    \sum_{k=1}^P \lambda_k^2
  }.
  \label{eq:effective_dimension}
\end{equation}
This metric, commonly used in statistical physics, quantifies the
number of significant principal components spanning the data manifold
in the feature space.

\begin{figure}[t]
  \centering
  \includegraphics[width=\columnwidth]{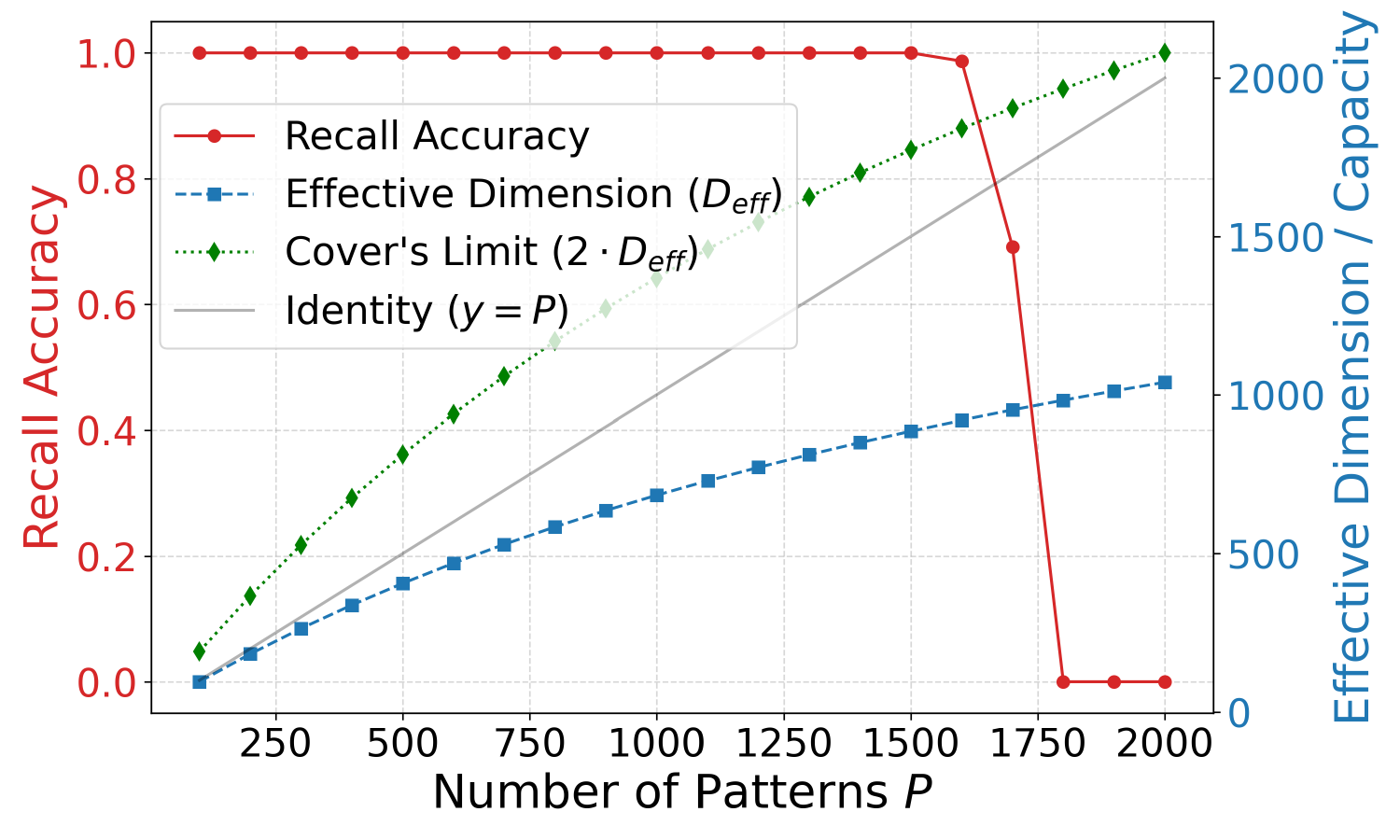}
  \caption{Geometric separability versus dynamic stability.
    Comparison between the actual storage limit (red) and a geometric
    reference point ($2D_{\text{eff}}(\bm{K})$, green) inspired by
    Cover's theorem.  The solid gray line ($y=P$) is shown as a
    reference corresponding to the identity function.}
  \label{fig:cover_theorem}
\end{figure}

Figure~\ref{fig:cover_theorem} compares the storage load $P$ with the
geometric reference point $2D_{\text{eff}}(\bm{K})$.  Memory recall
collapses (red curve drops) even when $P$ remains well below this
geometric bound (green curve, $P < 2D_{\text{eff}}(\bm{K})$).  This
discrepancy suggests an important distinction: the feature-space
geometry supports geometric separability beyond the observed capacity
limit, whereas the effective capacity is constrained by the stability
margin (SNR) required to maintain robust attractors.  The system fails
not because a separating solution does not exist geometrically, but
because the dynamics become unstable before reaching that structural
limit.

This dynamical instability can also be understood at the microscopic
level.  In a high-dimensional feature space, even if a hyperplane
exists that perfectly separates the patterns (satisfying Cover's
condition), the margin around this hyperplane may become arbitrarily
small as $P$ increases.  Consequently, a single step of the recurrent
dynamics perturbed by crosstalk noise can push the state vector across
the boundary into an incorrect basin.  These observations suggest
that, for recurrent associative memories, robust dynamical error
correction (high SNR) imposes a substantially stricter constraint than
static linear separability alone.

\section{Discussion}
\label{sec:discussion}

Our empirical and theoretical analyses, ranging from storage
experiments to geometric and statistical investigations of attractor
dynamics, suggest a characteristic self-organization mechanism in
KLR-trained Hopfield networks.  This section discusses the broader
implications of these findings for understanding high-capacity
associative memory.

\subsection{Exemplar-Based Memory with Localized Basins}

Our experiments with real-world data (Table~\ref{tab:cifar_capacity})
and the intra-class morphing analysis
(Fig.~\ref{fig:attractor_transition}(b)) suggest that the KLR network,
under high-load conditions, functions primarily as an
\textit{exemplar-based memory}~\cite{Nosofsky1986}.  Rather than
forming broad ``concept'' attractors that merge similar instances, the
network appears to assign a distinct, stable basin to each individual
stored pattern.

The highly localized, phase-transition-like boundaries observed
between these basins, supported by the critical slowing down
phenomenon (Fig.~\ref{fig:slowing_down}), allow the network to
separate a large number of patterns with minimal overlap.  We
hypothesize that this localized geometric partitioning of the state
space is an important factor underlying its high empirical storage
performance.

A useful comparison can be made with Modern Hopfield
Networks~\cite{Krotov2016, Ramsauer2021}, which explicitly induce
sharp attractor boundaries by introducing steep exponential or
polynomial nonlinearities into the energy function.  Our results
demonstrate that KLR networks achieve a comparably sharp,
phase-transition-like partitioning solely through the optimization of
synaptic weights, while retaining a standard quadratic energy
structure in the feature space.

\subsection{Geometric Separability and Dynamical Stability}
An important conclusion of this study is the distinction between
geometric separability and dynamical storage limits.  The analysis
inspired by Cover's theorem (Fig.~\ref{fig:cover_theorem}) indicates
that the feature space retains sufficient geometric dimensionality to
linearly separate more patterns than can be stably retrieved
($P < 2D_{\text{eff}}$).

However, the SNR analysis (Fig.~\ref{fig:snr_analysis}) demonstrates
that the primary limitation for retrieval is \textit{dynamical
  stability}.  Memory recall collapses not necessarily because a
geometric separating solution is no longer available, but because the
restoring signal strength becomes insufficient to overcome the
accumulating crosstalk noise during iterative retrieval dynamics.

These results suggest that, for robust associative memory, the
existence of a separating hyperplane is a necessary condition, whereas
robust dynamical error correction constitutes the dominant constraint.

\subsection{The Ridge as a Stable High-Capacity Regime}
The ``Ridge of Optimization'' can be understood in light of these
dynamical findings.  It represents a regime located near the
transition between two distinct behaviors: a stable phase in which
attractors are robustly maintained against noise, and an overloaded
phase in which crosstalk noise dominates and retrieval fails.

Our sequence memory experiments demonstrate that the network can
sustain stable limit cycles at very high loads ($P/N=16$), whereas a
marginal increase in load drives the system into the unstable regime.
This behavior, in which the system operates near the boundary between
stable and unstable dynamics, shares conceptual similarities with
systems operating near the \textit{edge of chaos}~\cite{Langton1990}
or exhibiting \textit{Self-Organized Criticality}~\cite{Beggs2003},
achieving high capacity while maintaining stability.

Although the present experiments primarily used the representative
value $\gamma=0.02$, our previous phase-diagram analysis
~\cite{tamamori_nolta_b} showed that the ``Ridge'' forms a continuous
operating region in hyperparameter space.  The sharp boundaries and
high capacity observed here are therefore expected to persist
throughout this regime, rather than arising from narrowly tuned
hyperparameters.

\subsection{Potential Implications for Retrieval Systems}
The mechanisms identified here may also have implications for
practical retrieval systems.  The exemplar-based nature of the memory,
combined with sharp decision boundaries and strong noise robustness,
shares several characteristics with large-scale, high-precision
retrieval modules, potentially providing a conceptual framework for
systems related to Retrieval-Augmented Generation
(RAG)~\cite{Lewis2020}.

Furthermore, the observation that the practical storage limit is
governed by SNR rather than pure geometric dimensionality suggests a
possible direction for algorithmic improvement.  Future learning rules
that explicitly minimize the variance of crosstalk interference may
further increase the effective storage load.

\subsection{Limitations and Future Work}
While this study provides a geometric and dynamical characterization
of KLR networks in specific regimes, several practical and theoretical
limitations remain.

From a computational perspective, the standard synchronous retrieval
process in kernel methods scales as $O(NP)$, which can become
computationally demanding for large-scale datasets.  To mitigate this
cost, future work should explore sparse kernel approximation
techniques, such as the Nystr\"{o}m method~\cite{Williams2000} and
random Fourier features~\cite{Rahimi2007}.  Recent work has also
demonstrated that this computational burden can be substantially
alleviated through ultra-low-precision weight
quantization~\cite{tamamori_2026} or by adopting an asynchronous
event-driven update scheme, which leverages large-margin attractors to
significantly reduce the number of required state
evaluations~\cite{tamamori_async_2026}.

Beyond computational considerations, our analysis primarily focused on
uncorrelated random patterns and specific image-embedding datasets.
How the network behaves on data with more complex hierarchical
correlation structures, or on temporal sequences with long-range
dependencies, remains an open question.

From a theoretical standpoint, our explanation of the capacity limit
relied on a simplified statistical SNR model.  A more rigorous
derivation using tools from statistical mechanics, such as replica
theory~\cite{Mezard1986, Amit1985} or random matrix
theory~\cite{Bai2009, Baik2005}, would be needed to obtain more
precise analytical characterizations of the phase boundaries.
Addressing these theoretical gaps and extending the geometric
framework to other learning rules remain important directions for
future work.

\section{Conclusion}
\label{sec:conclusion}
In this paper, we presented a systematic investigation of the
geometric and dynamical properties of high-capacity KLR-trained
Hopfield networks.  Beyond storage performance evaluation, we analyzed
the global structure of the attractor landscape and the mechanisms
governing the storage limit.

Our empirical evaluations showed that the network can maintain stable
limit cycles for random sequences up to a capacity of
$P/N \approx 16$, while robustly retrieving structured image data at
even higher effective loads.  Through detailed morphing experiments,
we showed that attractors in the optimal ``Ridge'' regime are
separated by highly localized, sharp boundaries.  The observation of
critical slowing down and steep effective potential barriers further
suggests that the network partitions the state space into distinct
exemplar-based basins with minimal ambiguous regions.

Our theoretical analysis provided a dynamical interpretation of the
eventual collapse of memory.  By comparing a statistical SNR analysis
with geometric separability considerations related to Cover's theorem,
we clarified that the practical storage limit is primarily constrained
by dynamical stability.  This limit corresponds to the point at which
restorative signal strength is overwhelmed by accumulating crosstalk
noise, rather than by geometric separability limits in the feature
space.

Overall, our findings characterize the Ridge of Optimization as a
high-capacity operating regime that maintains a necessary margin of
dynamical stability.  The interplay between geometric separability and
robust error correction identified here may provide a useful
perspective for the future development of large-scale associative
memory architectures.

\funding
Not applicable.

\conflictsofinterest
The author declares no competing interests.

\authorcontribution
The sole author contributed to the present work.

\aitools
The author used ChatGPT (GPT-5.3) and Gemini 2.5 Pro for proofreading
the English manuscript.

\end{document}